\documentclass[journal,twocolumns,twoside,print]{ieeecolor}
\usepackage{jsen}
\usepackage{cite}
\usepackage{amsmath,amssymb,amsfonts}
\usepackage{graphicx}
\usepackage{textcomp}
\usepackage{wrapfig}
\usepackage{booktabs}
\usepackage{array}
\usepackage{multirow}
\usepackage{calc}
\usepackage{float}
\usepackage[normalem]{ulem}
\graphicspath{{./}}
\def\BibTeX{{\rm B\kern-.05em{\sc i\kern-.025em b}\kern-.08em T\kern-.1667em\lower.7ex\hbox{E}\kern-.125emX}}
\definecolor{abstractbg}{rgb}{1,1,1}
\newcommand{\panelimage}[3]{%
\begin{minipage}[t]{#1}\centering\textbf{#2}\\[-0.2ex]\includegraphics[width=\linewidth]{#3}\end{minipage}}

\begin{document}
\title{Novel Ex-vivo Calf Brain Model with Integrated Sub-Skull Force Sensors to Access Simulated Neurosurgical Procedures}
\author{Hamad Binhammad*, Matheus Ballestero*, Mohammed Babgi, Seana Shaka, Nima Hemati, Rothaina Saeedi, Aiden Mazidi, Bianca Giglio, Rukun Dou, Houssem-Eddine Gueziri, Amir Hooshiar, and \\Rolando F. Del Maestro %
\thanks{* H. Binhammad and Matheus Ballestero are co-first authors.\\ H. Binhammad, M. Babgi, M. Ballestero, S. Shaka, B. Giglio, R. Dou, H.-E. Gueziri, and R. F. Del Maestro are with the Neurosurgical Simulation and Artificial Intelligence Learning Centre, Department of Neurology and Neurosurgery, Montreal Neurological Institute and Hospital, McGill University, Montreal, QC, Canada.}%
\thanks{H. Binhammad and M. Babgi are with the Department of Neurology and Neurosurgery, Montreal Neurological Institute and Hospital, McGill University, Montreal, QC, Canada.}%
\thanks{M. Babgi is with the Department of Neurosurgery, Ministry of the National Guard-Health Affairs, Jeddah, Saudi Arabia; the College of Medicine, King Saud Bin Abdulaziz University for Health Sciences, Jeddah, Saudi Arabia; and King Abdullah International Medical Research Center, Jeddah, Saudi Arabia.}%
\thanks{M. Ballestero is with the Department of Medicine, Federal University of Sao Carlos, Sao Carlos, Sao Paulo, Brazil.}%
\thanks{S. Shaka, R. Saeedi, and R. Dou are with the Faculty of Medicine and Health Sciences, McGill University, Montreal, QC, Canada.}%
\thanks{N. Hemati, A. Mazidi, and A. Hooshiar are with the Surgical Performance Enhancement and Robotics (SuPER) Centre, McGill University, Montreal, QC, Canada.}%
\thanks{R. Saeedi is with the Division of Pediatric Neurosurgery, Montreal Children's Hospital, McGill University, Montreal, QC, Canada, and the Department of Neurosurgery, Ministry of National Guard Health Affairs, King Abdullah Specialized Children Hospital, Jeddah, Saudi Arabia.}%
\thanks{B. Giglio and H.-E. Gueziri are with Dot-lab, Universite TELUQ, Montreal, QC, Canada.}%
\thanks{Corresponding author: M. Ballestero (e-mail: ballestero@ufscar.br).}}

\IEEEtitleabstractindextext{%
\fcolorbox{abstractbg}{abstractbg}{%
\begin{minipage}{\textwidth}%
\begin{wrapfigure}[16]{r}{3.6in}%
\includegraphics[width=3.5in]{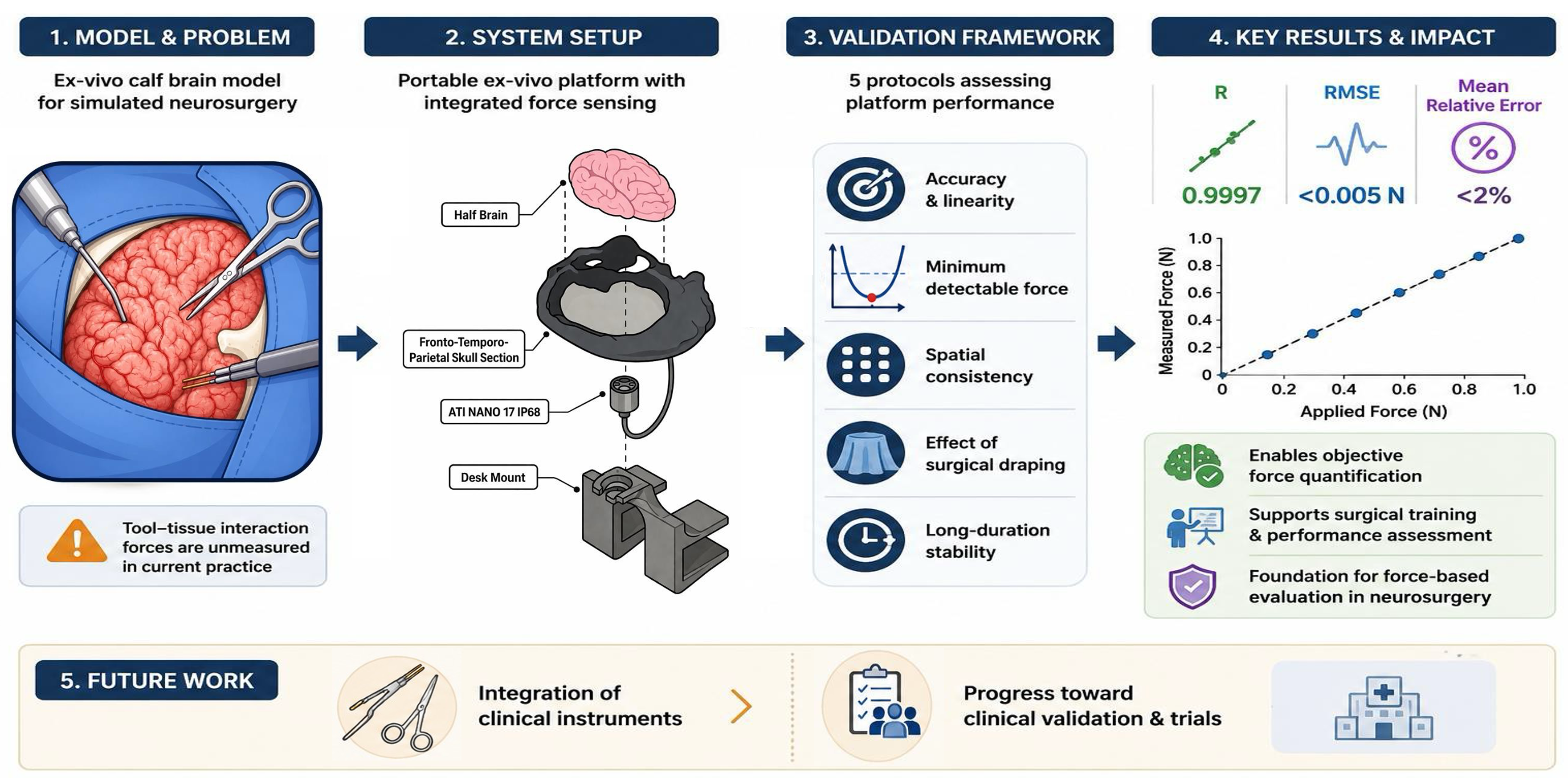}%
\end{wrapfigure}%
\begin{abstract}
Surgical tissue manipulation demands precision; however, tool-tissue manipulation force magnitudes under realistic conditions are rarely quantified. To address this gap, we proposed and validated a portable \textit{ex-vivo} force-sensing platform that measures tool-tissue interaction forces across the skull-brain interface during simulated neurosurgery. The system involves fresh calf brain tissue, used as a biological surrogate for brain parenchyma, placed in a 3D-printed human skull model equipped with a 6 degree-of-freedom force/torque sensor and a real-time data acquisition system. Five validation protocols assessed the accuracy and dynamic fidelity of the platform against ground-truth measurement, static accuracy and linearity using calibrated weights (0.5–50 g), minimum detectable force, spatial consistency across different anatomical regions, effect of surgical draping, and long-duration stability. Across protocols, measured forces showed excellent agreement with reference loads (correlation R = 0.9997), with RMSE $<$ 0.005 N and mean relative error under 2\%. The platform reliably detected low-magnitude forces down to 1 g (9.8 mN), while surgical drapes introduced no meaningful signal distortion and prolonged recordings exhibited minimal drift. Overall, the proposed framework provides objective, high-fidelity force quantification for skill training and performance assessment using fresh calf brain tissue and may serve as a foundation for force-based evaluation across other surgical procedures. Future work will integrate clinically used surgical instruments to increase procedural realism and will progress toward clinical trials to evaluate usability, educational impact, and translational relevance in practice-adjacent settings.
\end{abstract}

\end{minipage}}}
\maketitle
\section{Introduction}\label{introduction}
\subsection{Background}\label{background}

Precise force application is essential for the mastery of neurosurgical
procedures. Brain tissue has very limited tolerance to mechanical
stress, and small errors can result in tissue injury and patient
morbidity. During subpial resections of epileptic foci and intracranial
tumors, surgeons rely heavily on tactile feedback to distinguish
pathological from normal tissue and to modulate tissue deformation,
traction, and resection forces. These forces are rarely quantified in
the operating room, and surgeons are unaware of the exact magnitude of
force applied during operative maneuvers. Excessive, insufficient, and
poorly controlled force is associated with bleeding, tissue injury, and
adverse outcomes, whereas expert surgeons consistently demonstrate lower
and more stable force profiles than trainees \cite{ref1,ref2,ref3}.

Virtual reality (VR) simulators, such as the NeuroVR, have demonstrated
the feasibility for capturing force, trajectory, and efficiency metrics
during simulated brain tumor resections. Force-derived parameters,
including force bandwidth, cumulative applied force, and regional force
histogram, have been shown to differentiate expert neurosurgeons from
novices and quantify surgical safety, quality, and efficiency \cite{ref4},
establishing force as a clinically meaningful determinate of surgical
performance rather than a mechanical by-product of tissue interaction.
While simulators provide valuable environments for procedural repetition and cognitive learning, they often lack haptic realism and biological response dynamics \cite{ref5new}\cite{ref6new} \cite{ref7new}. To address these limitations,
\emph{ex-vivo} animal brain models have been developed to combine
biomechanical authenticity with controlled experimental measurement.
Winkler-Schwartz et al. \cite{ref8new} introduced an \emph{ex-vivo} calf brain
platform incorporating artificial tumors and motion tracking to
objectively assess surgical performance and extent of resection.
Almansouri et al. \cite{ref9new} have validated the platform's realism and
educational utility for training subpial corticectomy. However, neither
study incorporated direct force measurement into the \emph{ex-vivo}
platform, leaving a critical gap in the objective assessment of surgical
force during biologically realistic tissue interaction.

\subsection{Related Works}\label{related-works}

Efforts to quantify surgical forces have followed several approaches. A
systematic review of tool-tissue interaction forces across surgical
specialties found that brain tissue requires the lowest manipulation
forces (mean of 0.4 N) and that novices exert approximately 23\% more
force than experts \cite{ref10new}. Neurosurgery-specific studies have measured
both surgeon-tool contact forces and tool-tissue interaction forces on
bench models, reporting tool-tissue forces in the range of 0.01--0.59 N
during typical neurosurgical tasks \cite{ref11new}. To capture these forces in
real time, several groups have developed sensorized surgical
instruments. Force-sensing bipolar forceps have been used to quantify
forces during cadaveric neurosurgical dissection \cite{ref12new}, fiber Bragg
grating--based sensors have been integrated into aspiration instruments
for tridirectional force and torque detection \cite{ref13new},
three-dimensional force perception techniques have been proposed for
bipolar tips during tumor resection \cite{ref14new}, and smart haptic hand-held
devices have been designed for neurosurgical microdissection \cite{ref15new}.
In addition, recent work has explored miniaturized optical fiber--based
force sensing for minimally invasive surgery, including a learning-based
calibrated optical tactile sensor capable of measuring forces with a
mean absolute error as low as $\sim$0.1 N \cite{ref16new}, as well as
image-based optical fiber force sensing approaches that eliminate the
need for photodetectors and demonstrate reliable \emph{ex-vivo}
validation in biological tissue \cite{ref17new}. These approaches highlight the
growing trend toward compact, scalable, and learning-integrated sensing
solutions for surgical force quantification. While these
instrument-based approaches offer the advantage of measuring force
directly at the tool tip, they require modification of surgical
instruments, introduce additional costs, and are not widely available
for routine training use.

In parallel, robotic surgery has made notable advances in force sensing.
Robotic grippers with integrated force detection and feedback have been
developed for neurosurgical applications \cite{ref18new}, and robot-assisted
systems such as the neuroArm (Project neuroArm, University of Calgary,
Calgary, Canada) have enabled quantification of workspace and tool
forces during live neurosurgical procedures, reporting maximum forces of
1.86 N \cite{ref19new}. These robotic platforms typically achieve high
measurement precision but are designed for robot-assisted operative
workflows and are not directly applicable to manual surgical training or
assessment in non-robotic settings.

A key distinction of the proposed platform relative to prior work lies in the
physical location of the sensing element. Existing approaches can be grouped into
five categories. First, instrument-mounted sensors embed the transducer within the surgical tool itself. This includes force-sensing bipolar forceps applied
intraoperatively \cite{ref1,ref2,ref3,ref20new} or on bench and cadaveric models
\cite{ref11new,ref12new}, which have reported tool--tissue forces in the range of
0.01--1.84 N and have shown that novices exert significantly higher forces than
experienced surgeons, as well as a growing family of purpose-built transducers
based on fiber Bragg gratings, optical fibers, and capacitive elements integrated
into aspirators, forceps, and microdissectors \cite{ref13new,ref14new,ref15new,ref16new,ref17new,ref27new,ref21new}. These devices measure force
directly at the tool tip, but they require dedicated sensorized instrumentation,
add cost, and are therefore incompatible with the unmodified instruments used in
routine surgical training. Second, robot-integrated sensing embeds the transducer
in the manipulator rather than in a hand-held tool \cite{ref18new,ref19new}; the neuroArm system, for example, quantified tool forces up to 1.86 N during live procedures, yet such measurements are confined to robot-assisted workflows and cannot assess manual microsurgical technique. Third, virtual reality simulators
\cite{ref4,ref5new,ref6new} and machine-learning analyses derived from them
\cite{ref22new} have established force as a valid discriminator of expertise, with
classification accuracies as high as 90\%, but they operate exclusively on virtual tissue and cannot reproduce viscoelasticity, pial membranes, or subpial resection planes. Fourth, physical and \emph{ex vivo} simulation platforms provide biological or anatomical realism while capturing instrument kinematics, resection volume, or operator muscle activity rather than tool--tissue force
\cite{ref8new,ref9new,ref26new}; notably, existing \emph{ex vivo} calf brain simulators have demonstrated face and content validity for subpial corticectomy but incorporate no force measurement. Fifth, force-based skill assessment has been validated in other specialties using instrumented graspers, endoscopes, and robotic consoles \cite{ref23new,ref24new,ref25new}, confirming the educational value of force metrics while retaining the same instrument-dependency constraint. The proposed system addresses these limitations simultaneously by relocating the sensing element to the tissue side of the interaction: a 6-DOF force/torque sensor placed beneath a 3D-printed skull containing fresh calf brain tissue enables instrument-agnostic measurement of tool--tissue interaction forces on biological tissue, without any modification to the surgical instrument. Table~\ref{tab:comparison} summarizes these prior approaches and highlights the unique contributions of the proposed platform.

\begin{table*}[!p]
\caption{Comparison of Existing Force-Sensing Approaches in Surgical Simulation and Skill Assessment with the Proposed Platform}
\label{tab:comparison}
\centering
\scriptsize
\setlength{\tabcolsep}{3pt}
\renewcommand{\arraystretch}{1.1}
\begin{tabular}{p{0.085\textwidth}p{0.10\textwidth}p{0.115\textwidth}p{0.20\textwidth}p{0.20\textwidth}p{0.185\textwidth}}
\toprule
\textbf{Category} & \textbf{Study and Year} & \textbf{Application / Specialty} & \textbf{What They Used} & \textbf{Performance Metric} & \textbf{Key Limitation}\\
\midrule
I. Instrument-mounted sensors (clinical / cadaveric)
& Gan \emph{et al.}, 2015 \cite{ref3} & Neurosurgery, live operating room & Prototype strain-gauge force-sensing bipolar forceps & Dissection, coagulation and manipulation forces recorded intraoperatively & Pilot study; one bespoke instrument; no tissue-side reference\\
& Sugiyama \emph{et al.}, 2017 \cite{ref2} & Brain arteriovenous malformation surgery & Force-sensing bipolar forceps & Mean 0.23$\pm$0.06 N vs.\ 0.33$\pm$0.05 N (successful vs.\ unsuccessful) & Single pathology; forceps-specific; not used in training\\
& Sugiyama \emph{et al.}, 2018 \cite{ref20new} & Neurosurgery, skill assessment & Force-sensing bipolar forceps; normative force catalogue & Novices exerted significantly greater force than experts ($P<.001$) & Requires sensorized forceps in the OR; tool-specific\\
& Albakr \emph{et al.}, 2022 \cite{ref1} & Hemangioblastoma surgery & SmartForceps, strain gauges on each prong & 718 trials; mean force 0.20$\pm$0.17 N across tasks & Tool-specific; no ground-truth calibration against known loads\\
& Aggravi \emph{et al.}, 2016 \cite{ref11new} & Neurosurgery, bench model & Instrumented tool with hand- and tool-level sensors & Tool--tissue 0.01--0.59 N; surgeon--tool 0.01--6.6 N & Synthetic phantom; no biological tissue\\
& Maddahi \emph{et al.}, 2016 \cite{ref12new} & Neurosurgery, cadaveric dissection & Sensorized instruments, real-time force measurement & Peak dissection forces 0.50--1.84 N & Cadaveric tissue lacks turgor; instrument modification required\\
\cmidrule(l){2-6}
I. Instrument-mounted sensors (transducer development)
& Payne \emph{et al.}, 2015 \cite{ref15new} & Neurosurgical microdissection & Hand-held device with vibrotactile force-threshold alert & Threshold feedback reduced applied forces in user testing & Threshold only; no continuous quantification on tissue\\
& Li \emph{et al.}, 2019 \cite{ref13new} & Neurosurgery, aspiration & Disposable FBG-based tridirectional force/torque sensor & Decoupled tridirectional sensing at the aspirator tip & Requires FBG interrogator; one instrument type; bench only\\
& Bandari \emph{et al.}, 2019 \cite{ref16new} & Minimally invasive surgery & Optical force sensor, learning-based nonlinear calibration & MAE 0.085$\pm$0.096 N; linearity 96\%; fit 93\% & Error near the range of clinically relevant brain forces\\
& Bandari \emph{et al.}, 2020 \cite{ref17new} & Minimally invasive surgery & Image-based optical-fiber sensor without photodetectors & Agreement with ground truth in \emph{ex vivo} tissue & Non-neural tissue; no simulation environment\\
& Zhang \emph{et al.}, 2021 \cite{ref14new} & Brain tumor resection, robotic & 3D force perception at bipolar forceps tips & Range 0--4 N; reported precision $>$95\% & Robotic forceps only; no biological validation\\
& Liu \emph{et al.}, 2023 \cite{ref21new} & Robotic tissue manipulation & Haptics-enabled forceps, multimodal force sensing & Multimodal force estimation toward task autonomy & Custom forceps and robotic platform required\\
& Hussain \emph{et al.}, 2024 \cite{ref27new} & Robotic surgical systems & Decoupled multi-axis capacitive tactile sensor & Normal 0--10 N, shear 0--3.1 N; low repeatability error & Component-level only; no tissue or procedural validation\\
\midrule
II. Robot-integrated sensing
& Yoneyama \emph{et al.}, 2013 \cite{ref18new} & Neurosurgery, micro-manipulation & Force-detecting gripper with feedback system & Gripping and pulling forces resolved ($\sim$0.01 N resolution) & Gripper-specific; not applicable to standard instruments\\
& Maddahi \emph{et al.}, 2016 \cite{ref19new} & Robot-assisted neurosurgery & neuroArm telerobotic system, tool-tip force sensing & Maximum tool force 1.86 N; bandwidth $<$20 Hz & Restricted to robot-assisted workflow; no manual technique\\
\midrule
III. VR simulators with force metrics
& Azarnoush \emph{et al.}, 2015 \cite{ref4} & Simulated brain tumor resection & NeuroTouch/NeuroVR force pyramids and histograms & Force metrics discriminated psychomotor skill & Virtual tissue; no viscoelasticity or pial plane\\
& AlZhrani \emph{et al.}, 2015 \cite{ref5new} & Simulated brain tumor resection & NeuroTouch proficiency benchmarks & Experts resected less normal simulated tissue & Simulated haptics; no physical ground truth\\
& Bugdadi \emph{et al.}, 2018 \cite{ref6new} & Simulated brain tumor resection & NeuroVR force automaticity, Fitts and Posner model & Automaticity increased with expertise level & Virtual environment; no real tissue interaction\\
& Winkler-Schwartz \emph{et al.}, 2019 \cite{ref22new} & VR neurosurgical simulation & Machine learning on VR force and motion metrics & 90\% accuracy classifying four expertise levels & Trained exclusively on virtual-tissue data\\
\midrule
IV. Physical and \emph{ex vivo} platforms
& Winkler-Schwartz \emph{et al.}, 2020 \cite{ref8new} & Primary brain tumor neurosurgery & \emph{Ex vivo} calf brain, artificial tumors, instrument tracking & Extent of resection and kinematics quantified & No force measurement of any kind\\
& Almansouri \emph{et al.}, 2024 \cite{ref9new} & Subpial corticectomy training & \emph{Ex vivo} calf brain, continuous instrument tracking & Median face and content validity 6.0 of 7 & Kinematics only; tool--tissue force not captured\\
& Singh \emph{et al.}, 2023 \cite{ref26new} & Craniotomy simulation & 3D-printed skull, forearm force myography and ML & Skill classification accuracy 86.22$\pm$20.8\% & Measures operator muscle activity, not tool--tissue force\\
\midrule
V. Force-based assessment, other specialties
& Rahimi \emph{et al.}, 2023 \cite{ref25new} & Robot-assisted surgery & ForceSense tissue-handling force parameters & Force parameters discriminated skill levels & Console-dependent; synthetic tissue models\\
& Huang \emph{et al.}, 2024 \cite{ref23new} & Laparoscopic training & Intelligent grasper with real-time force feedback & Improved grip-force control; shortened learning curve & Instrument-embedded; not neurosurgical\\
& Bola \emph{et al.}, 2024 \cite{ref24new} & Rigid endoscopy & Force-sensor-instrumented model with feedback & Trainees reduced applied force after feedback & Non-biological manikin; single contact point\\
\midrule
\textbf{Proposed} & \textbf{This work} & \textbf{Simulated neurosurgery} & \textbf{Sub-skull 6-DOF force/torque sensor (ATI Nano17 IP68) beneath a 3D-printed hemi-skull with fresh \emph{ex vivo} calf brain; NI-6210 DAQ} & \textbf{R\,=\,0.9997; RMSE $<$0.005 N; MRE $<$2\%; minimum detectable force 1 g (9.8 mN); no drift over 10 min; unaffected by draping} & \textbf{Net transmitted force, not tool-tip force; no contact localization or multi-contact resolution; non-perfused ex vivo tissue}\\
\bottomrule
\end{tabular}
\end{table*}

\subsection{Objectives and
Contributions}\label{objectives-and-contributions}

While force metrics are known to correlate with surgical expertise and
safety, they have been primarily derived from VR-based environments
\cite{ref4}. As an alternative, \emph{ex-vivo} brain models exhibit superior
tissue fidelity but state-of-the-art still lacks direct tool-tissue
force measurement across the skull-brain interface. To address this gap,
we have designed, developed, and validated a novel \emph{ex-vivo} calf
brain model, integrated it with a sub-skull force sensing, and evaluated
its performance in simulated neurosurgical manipulation, i.e., subpial
resection. Our goal is to establish the accuracy, resolution,
repeatability, and clinical relevance of this force-sensing platform
through a series of structured verification protocols. This study
presents three contributions: (i) a novel portable system for
high-fidelity brain surgery simulation with tool-tissue force
measurement capability, (ii) verification of measurable forces during
subpial resections, and (iii) characterization of the minimum detectable
force and a comprehensive error analysis of the system as a measurement
instrument.

\section{Materials and Methods}\label{materials-and-methods}

\subsection{Experimental Setup}\label{experimental-setup}

The experimental setup was conceptualized as a surrogate for a realistic
surgical scenario. Fig.1 A-B depict the components and design of the
proposed platform. The reference measurements were acquired using an ATI Nano17
six-axis force/torque sensor (ATI Industrial Automation, Apex, NC, USA) with the
SI-12-0.12 calibration. Under this calibration, the sensor has a sensing range of
±12 N for Fx and Fy, ±17 N for Fz, and ±120 N·mm for Tx, Ty, and Tz. The corresponding resolutions with a 16-bit data acquisition system are 1/320 N ($\approx$3.1 mN) for all force axes and 1/64 N·mm ($\approx$15.6 mN·mm) for all torque axes. Per the manufacturer, the full-scale measurement uncertainty of the calibrated sensor is 0.75–1.25\%  of full scale, with repeatability better than 20\%  of the full-scale
measurement uncertainty under steady-state thermal conditions; the effects of
hysteresis and crosstalk are incorporated into the full-scale uncertainty
specification. The sensor further tolerates single-axis overloads of ±250 N (Fx,
Fy), ±480 N (Fz), ±1.6 N·m (Tx, Ty), and ±1.8 N·m (Tz).Additional components of the platform included a USB data acquisition board (NI-6210, National Instruments Inc., TX), a 3D-printed half skull simulating a fronto-temporo-parietal craniotomy prototyped from a publicly available CT scan (Embodi3D open-access medical imaging repository), and a 3D-printed desk mount. The 3D
reconstruction of the skull was performed in open-source 3D Slicer
v5.10.0 software. The desk mount was custom designed to receive the ATI
Nano17 sensor installed on the inferior surface of the skull. The mount
had a double-pronged custom press-fit feature to grip onto the edge of
lab desks, providing stability and security during experiments. The
parts were 3D-designed in Fusion 360 (v2605.1.52, Autodesk, CA). All the
parts were printed using Bambu Lab A1 (Bambu Lab, China) with
poly-lactic acid (PLA) filaments.

\begin{figure*}[!t]
\centering
\panelimage{1\textwidth}{}{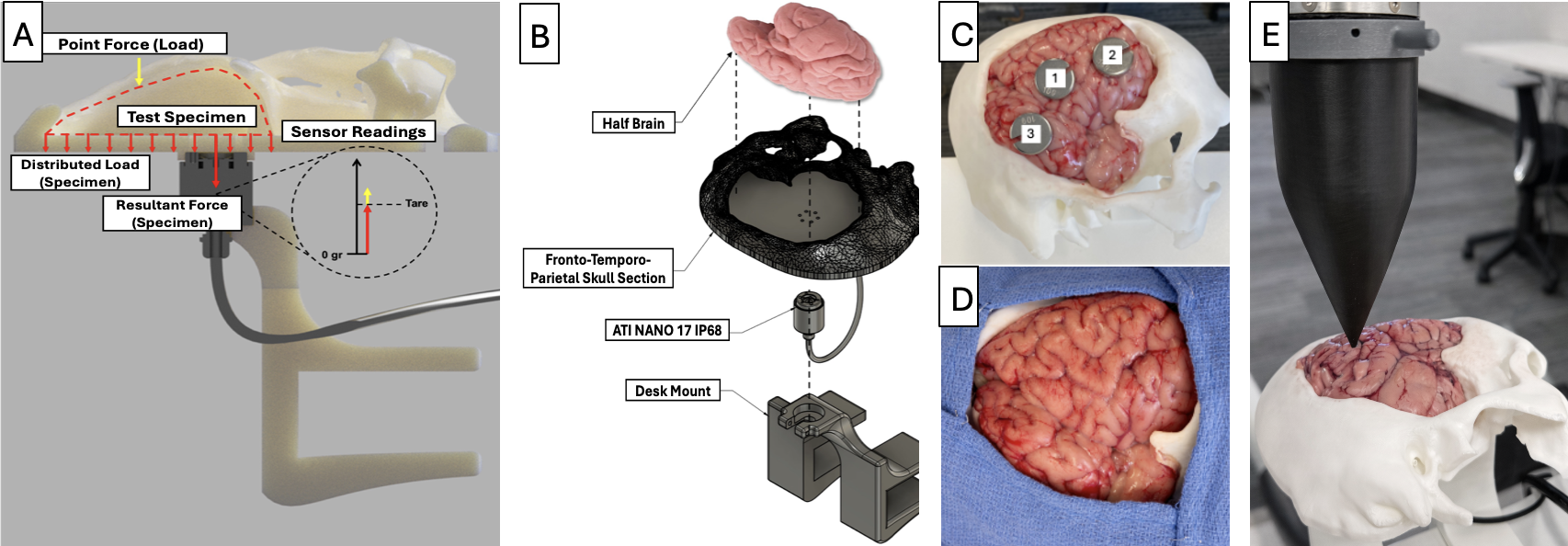}
\caption{Overview of the experimental setup: (A) cross-sectional view of the data acquisition setup showing the distribution of loads imparted on the specimen; (B) annotated components of the experimental setup; (C) fresh \emph{ex-vivo} calf brain mounted within a 3D-printed hemi-skull model with parietal (1), frontal (2), and temporal (3) anatomical locations; (D) model with surgical drapes; and (E) Kuka LBR iiwa 17 robotic arm mounted on the proposed model.}
\label{fig:setup}
\end{figure*}

\subsection{Validation Studies
Protocols}\label{validation-studies-protocols}

Fresh calf brains were obtained from a local butcher and transported under refrigeration. Prior to experimentation, brains were allowed to equilibrate to ambient room temperature of approximately 22°C to ensure consistent viscoelastic tissue properties across all testing sessions. All experiments were conducted within a controlled post-mortem time window (24h) to minimize the potential effects of tissue degradation on mechanical properties.	

Standardized weights
(0.5-50 g, McMaster-Carr, IL), surgical drapes, and a Kuka LBR iiwa 17
robotic arm with an integrated force/torque sensor were used to support
force application during the experimental protocols.

Five protocols were conducted to assess the performance, accuracy, and
clinical relevance of the platform. Specifically, we evaluated: the
fidelity of force sensing (Protocol 1), spatial consistency (Protocol
2), sensor resolution (Protocol 3), influence of surgical draping
(Protocol 4), and force sensing stability (Protocol 5). The experiments
collectively aimed to ensure that the system could consistently and
accurately measure a range of forces that compare to realistic
tool-tissue interactions in a neurosurgical context. Each protocol was
performed at the three predetermined spatially distributed locations, as
labelled in Fig. 1C.

\subsection{Protocol 1: Robotic Sub-Skull Dual Force Sensor}\label{protocol-1-robotic-sub-skull-dual-force-sensor}

The Kuka robotic arm executed controlled, repeatable tool trajectories
to apply consistent and variable force profiles, seen in Fig.~\ref{fig:setup}E. It was
programmed to produce a 2.5 mm downward displacement at three velocities
(0.04, 0.08, and 0.16 m/s) at each predefined brain location, repeated
as sinusoidal waveforms to deliver a range of controlled force profiles.
Interaction forces between the tool and tissue were measured with two
force/torque sensors: one native integrated in the robot end-effector
and one beneath the skull model.

\subsection{Protocol 2: Sensor Accuracy Validation Using Calibration
Weights}\label{protocol-2-sensor-accuracy-validation-using-calibration-weights}

The protocol evaluated the sub-skull sensor's linearity and accuracy by
applying known static reference loads. Standard 10 g (0.098 N) weights
were added every 10 seconds up to 50 g (0.49 N), then removed in reverse
order with the same timing. This manual loading-unloading sequence was
performed at three anatomical brain locations, each tested in three
trials, and replicated on a second calf brain, for a total of 18 trials
(n = 18). Measured forces were compared with applied loads to assess
accuracy, linearity, and consistency across spatially heterogeneous
tissue.

\subsection{Protocol 3: Minimum Detectable Force Resolution and Spatial
Consistency
Assessment}\label{protocol-3-minimum-detectable-force-resolution-and-spatial-consistency-assessment}

This protocol, divided into two parts, focused on verifying the minimum
detectable force as well as consistency across the brain surface.

Protocol 3A: Weights of 10, 5, 2, 1, and 0.5 g (total 18.5 g; 0.18 N)
were applied sequentially at 10-second intervals, then removed in
reverse order with the same timing. This loading-unloading sequence was
performed at three brain locations, each tested in three trials, and the
entire protocol was repeated on a second calf brain (18 trials total, n
= 18). The goal was to determine the smallest force that the sensor
could reliably measure in soft tissue even after exposure to higher
forces, thereby establishing the system's resolution threshold.

Protocol 3B: Using the lowest force magnitude defined in protocol 3A, we
conducted an additional protocol to assess spatial consistency of sensor
measurements across the brain surface. A single standardized 1 g weight
was applied to and removed from each of three predetermined locations at
fixed 10-second intervals. This sequence constituted one trial and was
repeated three times to evaluate reproducibility under minimal loading.
The protocol was then repeated on a second calf brain, for a total of 6
trials (n = 6).

\subsection{Protocol 4: Effect of Surgical Drapes on Force Measurement
Fidelity}\label{protocol-4-effect-of-surgical-drapes-on-force-measurement-fidelity}

This protocol tested whether standard surgical draping affected force
transmission and sensor readings. The setup and weight sequence matched
Protocol 2, except a standard surgical drape was placed over the skull
and exposed brain to simulate neurosurgical conditions (Fig.~\ref{fig:setup}D). The
protocol was run once at each of three anatomical locations on two brain
specimens, for six total trials (n = 6).

\subsection{Protocol 5: Stability of Recording Over Longer
Timeframe}\label{protocol-5-stability-of-recording-over-longer-timeframe}

This protocol assessed potential signal drifts during prolonged
recording under repeated loading-unloading cycles. The setup and loading
sequence matched Protocol 2, but the recording was extended to 10
minutes with continuous loading-unloading at 10-second intervals. Six
loading-unloading cycles were completed in this period. The protocol was
performed once at each of three anatomical locations (3 trials, n = 3).

\subsection{Error Analysis}\label{error-analysis}

We computed multiple error metrics to capture complementary dimensions
of performance: 1) root mean squared error (RMSE), 2) mean absolute
error (MAE), 3) maximum absolute error (MaxAE) to quantify the
worst-case instantaneous deviations, 4) mean relative error (MRE), and
5) maximum relative error (MaxRE) to quantify the worst proportional
mismatch. To avoid artifacts introduced during manual placement and
removal of calibration weights, force signals were segmented and only
steady-state intervals corresponding to stable load conditions were
analyzed. More specifically, we considered the middle 80\% of the data
in each segment of the experiments and excluded the first 10\% and last
10\% segments from error analysis. In addition, transient peaks (less
than 2 seconds in duration) associated with manipulation events were
excluded through manual extraction of the stable force plateaus prior to
error calculation.

\section{Results}\label{results}

\subsection{Protocol 1: Robotic Sub-Skull Dual Force
Sensor}\label{protocol-1-robotic-sub-skull-dual-force-sensor-1}

Across all three locations, the mean absolute error (MAE) between the
platform measurement and robot was below 1.3 mN, with values of 1.244
mN, 1.278 mN, and 0.993 mN for parietal, frontal, and temporal regions,
respectively, yielding a cross-location mean of 1.172 mN. The root mean
squared error (RMSE) followed a consistent trend, ranging from 1.657 mN
(temporal) to 2.142 mN (frontal), with an overall mean of 1.966 mN. The
error distributions, shown as inset histograms in Fig. 2, confirm a
zero-centered, approximately symmetric profile across all locations,
suggesting the absence of systematic bias in the Jacobian-based force
estimate. Measured forces closely match the theoretical values,
demonstrating high accuracy of the robotic force application system,
with negligible mean differences across all trials, as shown in Table~\ref{tab:protocol1}.

\begin{figure*}[!t]
\centering
\includegraphics[width=0.95\textwidth]{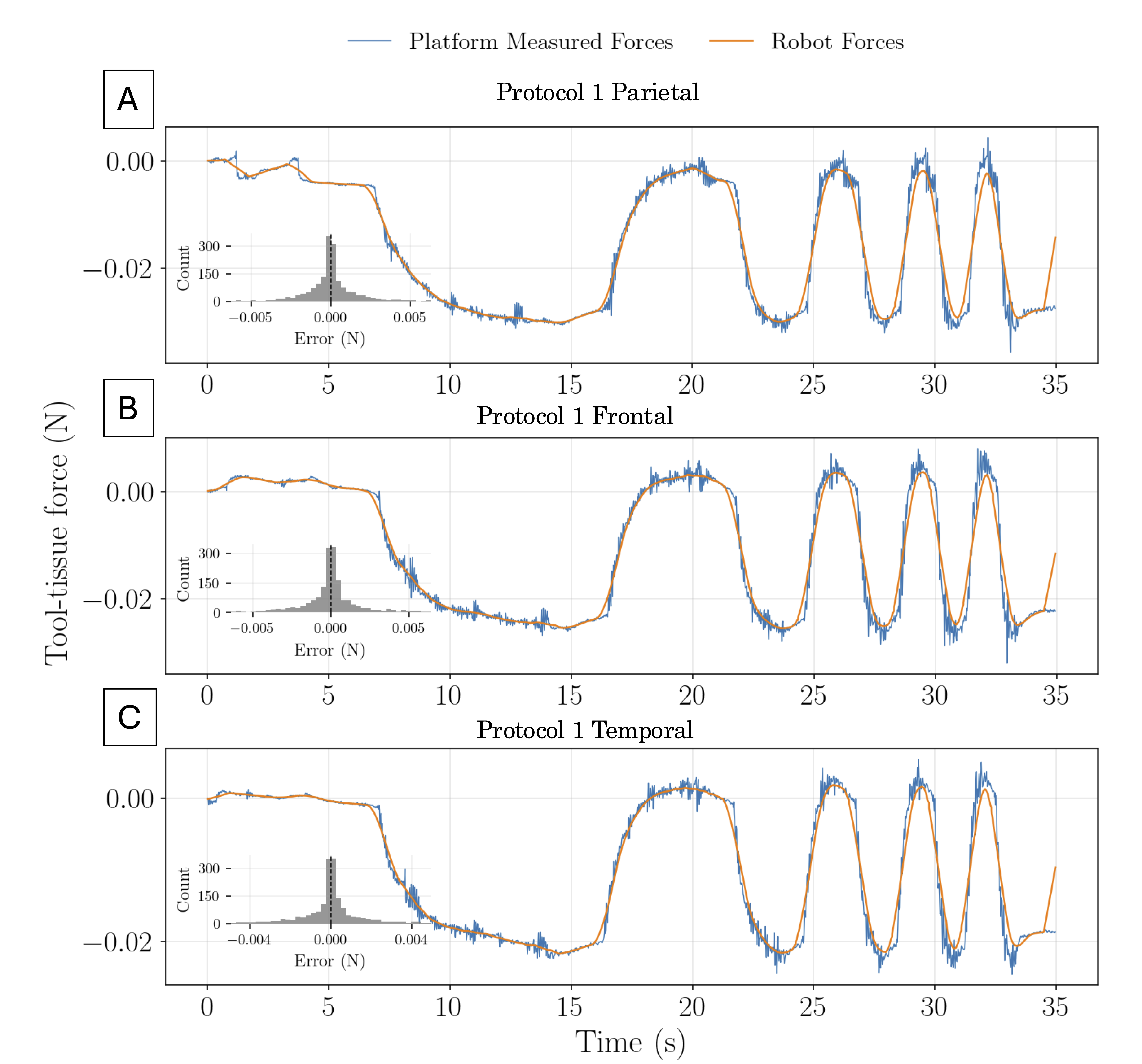}
\caption{Comparison of forces measured by the robot and the proposed platform in a chirp indentation test at (A) parietal, (B) frontal, and (C) temporal locations. Insets show histograms of error distributions centered around zero across all regions.}
\label{fig:protocol1}
\end{figure*}

\begin{table}[!t]
\caption{Protocol 1. Comparison of Robotically Applied and Subskull-Measured Forces}
\label{tab:protocol1}
\centering
\footnotesize
\begin{tabular}{>{\centering\arraybackslash}p{0.2\linewidth}>{\centering\arraybackslash}p{0.2\linewidth}>{\centering\arraybackslash}p{0.2\linewidth}>{\centering\arraybackslash}p{0.2\linewidth}}
\toprule
\textbf{Nominal Weight (g)} & \textbf{Theoretical Force (N)} & \textbf{Mean Measured Force (N)} & \textbf{Mean Difference (N)}\\
\midrule
50 & 0.49 & 0.49 & 0.00\\
100 & 0.98 & 0.98 & 0.00\\
150 & 1.47 & 1.47 & 0.00\\
200 & 1.96 & 1.96 & 0.00\\
250 & 2.45 & 2.45 & 0.00\\
300 & 2.94 & 2.94 & 0.00\\
\bottomrule
\end{tabular}
\end{table}

\subsection{Protocol 2: Sensor Accuracy Validation Using Calibration
Weights}

Results from the frontal, parietal, and temporal regions of Protocol 2
are shown in Fig. 3A-C. The sensor showed a clear stepwise response to
each incremental weight. Measured forces closely matched theoretical
values from the calibrated weights, confirming accuracy within the
tested range, as shown in Table~\ref{tab:protocol2}. Identical loading and unloading
profiles in all 18 trials demonstrated high repeatability and spatial
consistency.

\begin{table}[!t]
\caption{Protocol 2. Subskull Sensor Force Measurements under Incremental Loading and Unloading}
\label{tab:protocol2}
\centering
\footnotesize
\begin{tabular}{>{\centering\arraybackslash}p{0.2\linewidth}>{\centering\arraybackslash}p{0.2\linewidth}>{\centering\arraybackslash}p{0.2\linewidth}>{\centering\arraybackslash}p{0.2\linewidth}}
\toprule
\textbf{Nominal Weight (g)} & \textbf{Theoretical Force (N)} & \textbf{Mean Measured Force (N)} & \textbf{Mean Difference (N)}\\
\midrule
0 & 0.000 & 0.000 & 0.000\\
10 & 0.098 & 0.100 & +0.002\\
20 & 0.196 & 0.199 & +0.003\\
30 & 0.294 & 0.293 & -0.001\\
40 & 0.392 & 0.391 & -0.001\\
50 & 0.490 & 0.490 & 0.000\\
40 & 0.392 & 0.391 & -0.001\\
30 & 0.294 & 0.296 & +0.002\\
20 & 0.196 & 0.200 & +0.004\\
10 & 0.098 & 0.100 & +0.002\\
0 & 0.000 & 0.000 & 0.000\\
\bottomrule
\end{tabular}
\end{table}

\subsection{Protocol 3: Minimum Detectable Force Resolution and
Spatial Consistency
Assessment}\label{protocol-3-minimum-detectable-force-resolution-and-spatial-consistency-assessment-1}

Force recordings obtained during Protocol 3A demonstrated that the
sensing platform reliably detected very small loads, including forces as
low as 1 g, with clear and reproducible signal responses as shown in Fig
3D-F.

Using the minimum reliably detectable force identified in Protocol 3A,
measurements obtained during Protocol 3B demonstrated consistent force
detection across different anatomical locations, illustrated by Fig.
3G-I. Sequential application of the same small reference load at
multiple sites produced highly similar force readings, with minimal
variability between locations and excellent repeatability across trials.

Tables~\ref{tab:protocol3a} and~\ref{tab:protocol3b} present the performance of the sub-skull force sensor
under low-force and near-zero loading conditions. In Protocol 3A,
measured forces closely aligned with theoretical values, with minimal
deviations, demonstrating the sensor's ability to accurately resolve
small force magnitudes. Under near-zero loading in Protocol 3B, the
sensor exhibited a stable baseline with low noise and negligible drift,
indicating high sensitivity and measurement reliability. These findings
collectively highlight the sensor's robustness across low-force regimes
and its capability to detect subtle force changes with high precision.

\subsection{Protocol 4: Effect of Surgical Drapes on Force
Measurement
Fidelity}\label{protocol-4-effect-of-surgical-drapes-on-force-measurement-fidelity-1}
\normalcolor
Force measurements recorded in the presence of standard surgical draping
demonstrated signal profiles that were comparable to those obtained
under undraped conditions. Across all tested locations and specimens,
applied loads produced consistent and reproducible force responses
without evidence of significant damping, distortion, or measurement
artifacts attributable to the drape material, seen in Fig. 3J-L. Table~\ref{tab:protocol4} demonstrates consistent sensor performance and minimal deviation despite
the presence of surgical drapes.

\begin{table}[!t]
\caption{Protocol 3A. Subskull Sensor Response under Low-Force Incremental Loading}
\label{tab:protocol3a}
\centering
\footnotesize
\begin{tabular}{>{\centering\arraybackslash}p{0.2\linewidth}>{\centering\arraybackslash}p{0.2\linewidth}>{\centering\arraybackslash}p{0.2\linewidth}>{\centering\arraybackslash}p{0.2\linewidth}}
\toprule
\textbf{Nominal Weight (g)} & \textbf{Theoretical Force (N)} & \textbf{Mean Measured Force (N)} & \textbf{Mean Difference (N)}\\
\midrule
0 & 0.000 & 0.000 & 0.000\\
10 & 0.098 & 0.100 & +0.002\\
15 & 0.147 & 0.150 & +0.003\\
17 & 0.167 & 0.170 & +0.003\\
18 & 0.177 & 0.180 & +0.003\\
18.5 & 0.182 & 0.180 & +0.003\\
18 & 0.177 & 0.180 & +0.003\\
17 & 0.167 & 0.170 & +0.003\\
15 & 0.147 & 0.150 & +0.003\\
10 & 0.098 & 0.100 & +0.002\\
0 & 0.000 & 0.000 & 0.000\\
\bottomrule
\end{tabular}
\end{table}

\begin{table}[!t]
\caption{Protocol 3B. Subskull Sensor Response at 1 g Force Levels}
\label{tab:protocol3b}
\centering
\footnotesize
\begin{tabular}{>{\centering\arraybackslash}p{0.2\linewidth}>{\centering\arraybackslash}p{0.2\linewidth}>{\centering\arraybackslash}p{0.2\linewidth}>{\centering\arraybackslash}p{0.2\linewidth}}
\toprule
\textbf{Nominal Weight (g)} & \textbf{Theoretical Force (N)} & \textbf{Mean Measured Force (N)} & \textbf{Mean Difference (N)}\\
\midrule
0 & 0.000 & 0.000 & 0.000\\
1 & 0.0098 & 0.010 & +0.0002\\
0 & 0.000 & 0.000 & 0.000\\
1 & 0.0098 & 0.010 & +0.0002\\
0 & 0.000 & 0.000 & 0.000\\
1 & 0.0098 & 0.010 & +0.0002\\
0 & 0.000 & 0.000 & 0.000\\
\bottomrule
\end{tabular}
\end{table}

\begin{table}[!t]
\caption{Protocol 4. Subskull Sensor Force Measurements under Incremental Loading with Surgical Drapes}
\label{tab:protocol4}
\centering
\footnotesize
\begin{tabular}{>{\centering\arraybackslash}p{0.2\linewidth}>{\centering\arraybackslash}p{0.2\linewidth}>{\centering\arraybackslash}p{0.2\linewidth}>{\centering\arraybackslash}p{0.2\linewidth}}
\toprule
\textbf{Nominal Weight (g)} & \textbf{Theoretical Force (N)} & \textbf{Mean Measured Force (N)} & \textbf{Mean Difference (N)}\\
\midrule
0 & 0.000 & 0.000 & 0.000\\
10 & 0.098 & 0.099 & +0.001\\
20 & 0.196 & 0.200 & +0.004\\
30 & 0.294 & 0.297 & +0.003\\
40 & 0.392 & 0.396 & +0.004\\
50 & 0.490 & 0.490 & 0.000\\
40 & 0.392 & 0.397 & +0.005\\
30 & 0.294 & 0.297 & +0.003\\
20 & 0.196 & 0.200 & +0.004\\
10 & 0.098 & 0.100 & +0.002\\
0 & 0.000 & 0.000 & 0.000\\
\bottomrule
\end{tabular}
\end{table}

\subsection{Protocol 5: Stability of Recording Over Longer
Timeframe}\label{protocol-5-stability-of-recording-over-longer-timeframe-1}
\normalcolor
Continuous recordings obtained during repeated loading-unloading cycles
demonstrated stable sensor performance over extended acquisition
periods. Force measurements remained consistent across successive
cycles, with no progressive baseline drift, signal degradation, or loss
of responsiveness observed during prolonged recording, as shown in Fig.
3M-O. Table~\ref{tab:protocol5} compares theoretical and sub-skull--measured forces at
trough (0 g) and peak (50 g) points across six loading-unloading cycles,
demonstrating high repeatability and negligible drift over repeated
trials.

\begin{table}[!t]
\caption{Protocol 5. Sub-Skull Sensor Repeatability under Cyclic Loading}
\label{tab:protocol5}
\centering
\footnotesize
\begin{tabular}{>{\centering\arraybackslash}p{0.2\linewidth}>{\centering\arraybackslash}p{0.2\linewidth}>{\centering\arraybackslash}p{0.2\linewidth}>{\centering\arraybackslash}p{0.2\linewidth}}
\toprule
\textbf{Nominal Weight (g)} & \textbf{Theoretical Force (N)} & \textbf{Mean Measured Force (N)} & \textbf{Mean Difference (N)}\\
\midrule
0 & 0.000 & 0.000 & 0.000\\
50 & 0.490 & 0.490 & 0.000\\
0 & 0.000 & 0.000 & 0.000\\
50 & 0.490 & 0.490 & 0.000\\
0 & 0.000 & 0.000 & 0.000\\
50 & 0.490 & 0.490 & 0.000\\
0 & 0.000 & 0.000 & 0.000\\
50 & 0.490 & 0.490 & 0.000\\
0 & 0.000 & 0.000 & 0.000\\
50 & 0.490 & 0.490 & 0.000\\
0 & 0.000 & 0.000 & 0.000\\
\bottomrule
\end{tabular}
\end{table}

\subsection{Error Analysis}
\normalcolor
The protocols measured forces exhibited high levels of agreement with
the reference data. As seen in Fig. 4 and 5 and Table~\ref{tab:error_all}, each protocol showed similar performance, with mean relative errors under 2\% of
ground truth and RMSE consistently below 0.005 N.

The force comparisons between Protocols 2 and 4 were very similar, with
the RMSE, MRE, and MAE either exactly or near identical. The higher
loading increased error spread, whereas Protocol 3A showed a larger
standard deviation ($\sim$0.015 N vs. $\sim$0.0028 N)
than Protocols 2 and 4. This is consistent with the magnitudes of force
applied in each protocol.

Fig. 6 and Table~\ref{tab:error_sample} demonstrate the linear correlation between the measured signals and the reference weight application across the
protocols.

\begin{table}[!t]
\caption{Summary table of error metrics across all trials.}
\label{tab:error_all}
\centering
\footnotesize
\begin{tabular}{>{\centering\arraybackslash}p{0.2\linewidth}>{\centering\arraybackslash}p{0.2\linewidth}>{\centering\arraybackslash}p{0.2\linewidth}>{\centering\arraybackslash}p{0.2\linewidth}}
\toprule
\textbf{Metric} & \textbf{Protocol 2} & \textbf{Protocol 3A} & \textbf{Protocol 4}\\
\midrule
RMSE (N) & 0.0049 N & 0.0028 N & 0.0047 N\\
MRE (\%) & 1.70\% & 1.81\% & 1.67\%\\
MaxAE (N) & 0.0095 N & 0.0034 N & 0.0095 N\\
STD (N) & 0.0028 & 0.0015 & 0.0029\\
\bottomrule
\end{tabular}
\end{table}

\begin{table}[!t]
\caption{Summary table of error metrics within sample protocols.}
\label{tab:error_sample}
\centering
\footnotesize
\begin{tabular}{>{\centering\arraybackslash}p{0.2\linewidth}>{\centering\arraybackslash}p{0.2\linewidth}>{\centering\arraybackslash}p{0.2\linewidth}>{\centering\arraybackslash}p{0.2\linewidth}}
\toprule
\textbf{Metric} & \textbf{Protocol 2} & \textbf{Protocol 3A} & \textbf{Protocol 4}\\
\midrule
RMSE (N) & 0.0049 N & 0.0028 N & 0.0047 N\\
MRE (\%) & 1.70\% & 1.81\% & 1.67\%\\
MaxAE (N) & 0.0095 N & 0.0034 N & 0.0095 N\\
STD (N) & 0.0028 & 0.0015 & 0.0029\\
\bottomrule
\end{tabular}
\end{table}

\begin{figure*}[!t]
\centering
\panelimage{\textwidth}{}{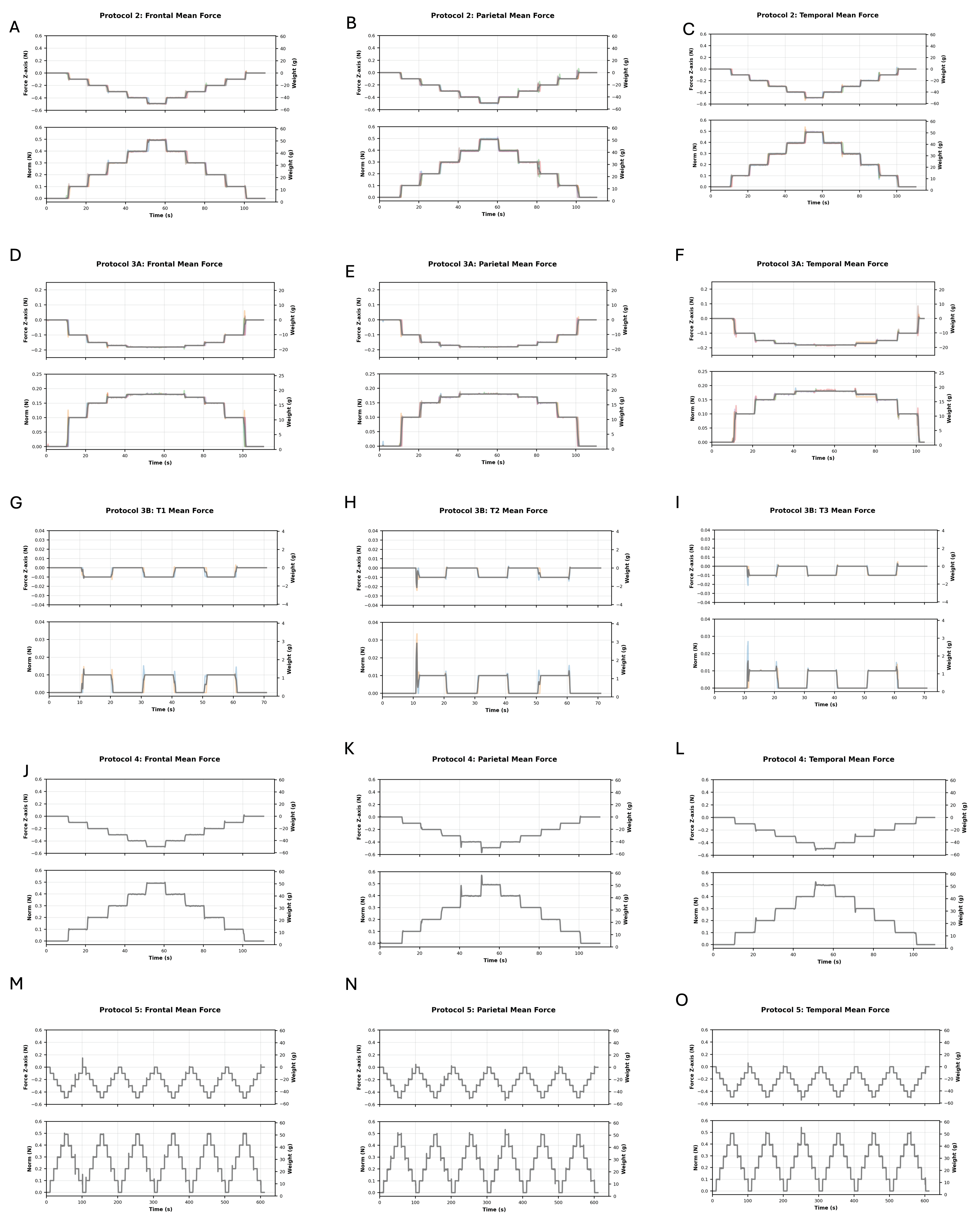}
\caption{Force measurements obtained across all experimental protocols at three anatomical locations: frontal (left column -- A, D, G, J, M), parietal (middle column -- B, E, H, K, N), and temporal (right column -- C, F, I, L, O). For each panel, the upper plots depict the vertical force component ($F_z$) transmitted to the brain surface, whereas the lower plots show the total force magnitude (norm of $F_x$, $F_y$, and $F_z$). Panels A--C correspond to Protocol 2, showing mean force profiles across six trials per location (grey traces represent the averaged response). Panels D--F correspond to Protocol 3A, also showing mean responses across six trials at each location. Panels G--I represent Protocol 3B, where a single 1 g load was sequentially applied at each location; each plot represents the mean of three trials per site. Panels J--L correspond to Protocol 4, evaluating the effect of surgical draping, with each plot representing the mean of two trials per location. Panels M--O correspond to Protocol 5, illustrating individual trial recordings during repeated loading cycles used to assess long-term signal stability and drift.}
\label{fig:all_protocols}
\end{figure*}

\begin{figure*}[!t]
\centering
\panelimage{0.75\textwidth}{}{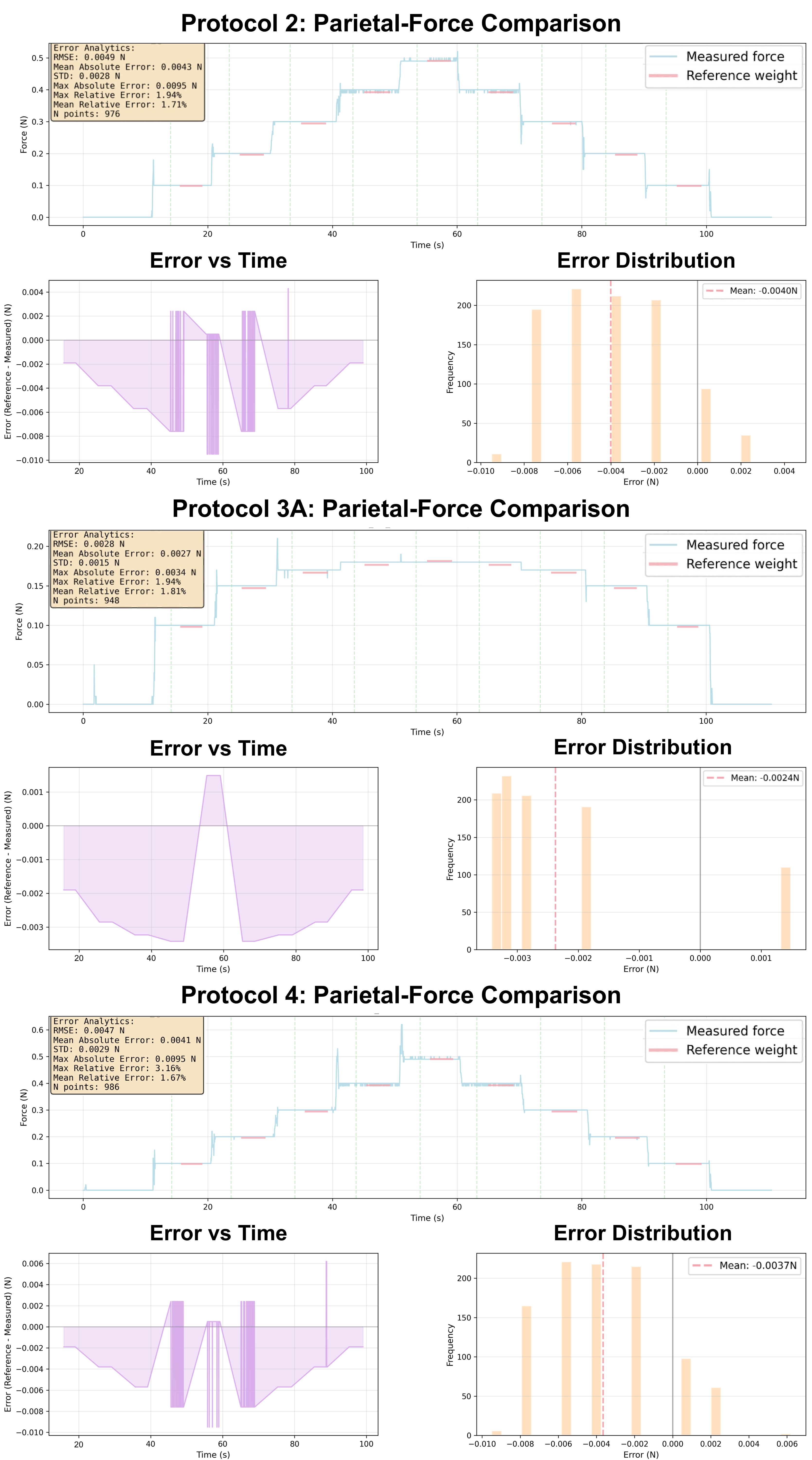}
\caption{Error distribution and absolute error of the measured forces in Protocols 2 (A), 3A (B), and 4 (C). The top panel shows the test period (red dots) used to compare the measured forces with the ground truth (calibrated weights). Abbreviations: RMSE, root mean square error; MAE, mean absolute error; STD, standard deviation; N, number of data points.}
\label{fig:error_distribution_protocols}
\end{figure*}

\begin{figure*}[!t]
\centering
\includegraphics[width=\textwidth]{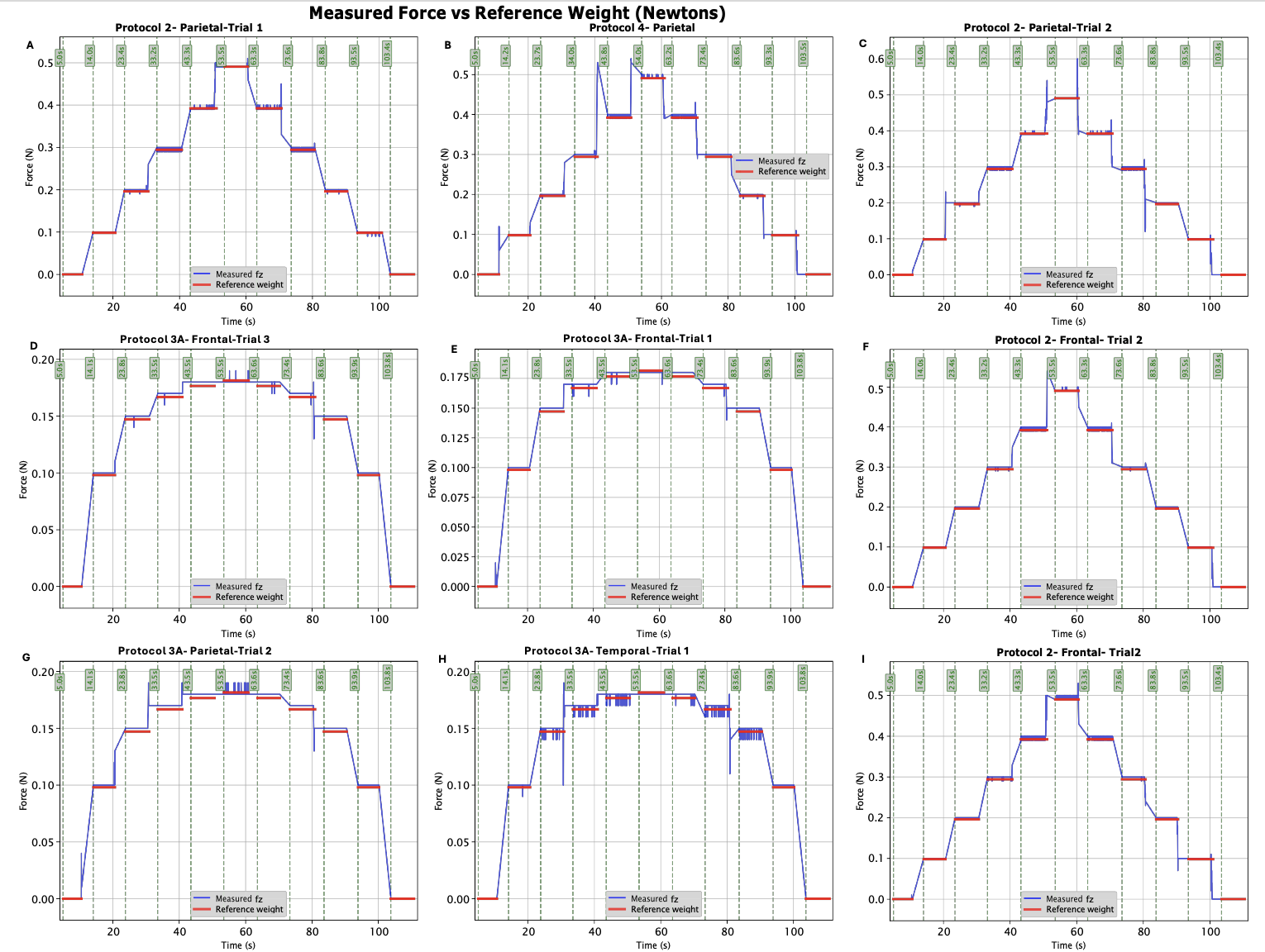}
\caption{Representative timestamps of a select group of trials with reference sampling period (red) and measured force during the trials (blue). Data in the red periods were used in the error analysis. (A) Protocol 2, parietal trial 1; (B) Protocol 4, parietal; (C) Protocol 2, parietal trial 2; (D) Protocol 3A, frontal trial 3; (E) Protocol 3A, parietal trial 1; (F) Protocol 2, frontal trial 2; (G) Protocol 3A, parietal trial 2; (H) Protocol 3A, temporal trial; and (I) Protocol 2, frontal trial 2.}
\label{fig:representative_timestamps}
\end{figure*}

\begin{figure*}[!t]
\centering
\panelimage{0.58\textwidth}{}{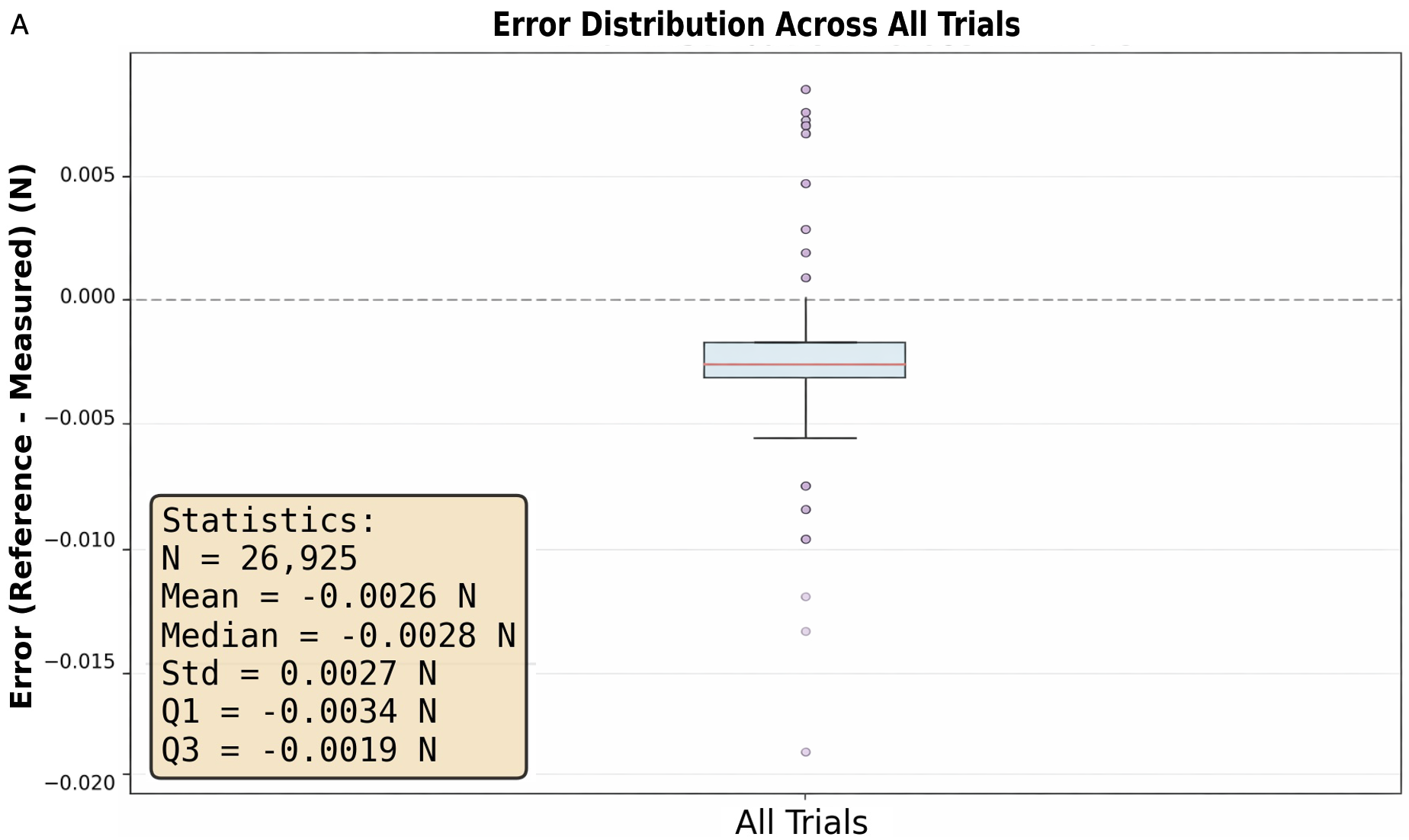}
\panelimage{0.4\textwidth}{}{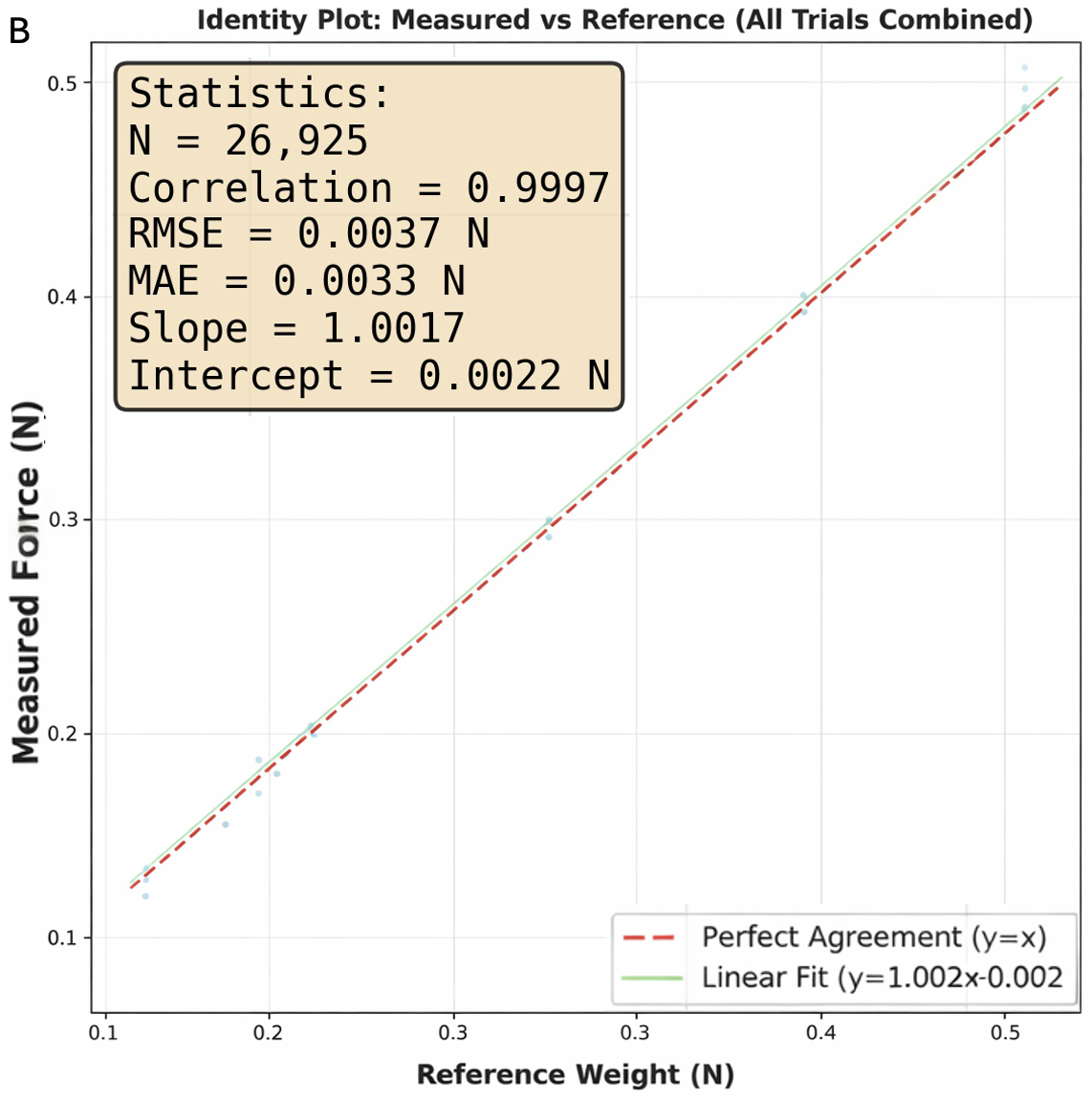}
\caption{(A) Error distribution across all trials. (B) Identity plot of measured vs. reference values (all trials). Abbreviations: N, number of data points; STD, standard deviation; Q1, first quartile; Q3, third quartile; RMSE, root mean squared error; MAE, mean absolute error.}
\label{fig:all_trials}
\end{figure*}

\section{Discussion}\label{discussion}

In neurosurgical procedures, surgical instrument force application is
critically important \cite{ref3}. The quantity and quality of force
application can significantly influence surgical outcomes \cite{ref20new}.
There is a lack of studies addressing the impact of surgical force on
the brain and how this force can be effectively measured and analyzed.
Furthermore, the absence of haptic feedback in VR-based surgical
simulation systems deprives surgeons of critical information normally
obtained through tissue contact \cite{ref2}. Previous studies have shown
that haptic perception can reduce operative time, facilitate surgical
training, enhance accuracy, and ultimately improve patient safety
\cite{ref3}. Consequently, a force-sensing mechanism is an essential
component of any haptic interface. However, the use of force monitoring
in neurosurgical training remains underappreciated, and neurosurgical
education continues to rely heavily on subjective assessment and
feedback from mentors \cite{ref2,ref3}.

In this study, we developed a novel proof-of-concept \emph{ex-vivo} calf
brain model for neurosurgical training with sub-skull force-sensing
capabilities, providing a platform that can be extended to a wide range
of other surgical procedures. Building on this framework, we implemented
structured validation protocols to characterize the platform's accuracy,
resolution, repeatability, and clinical relevance. By directly measuring
forces applied to biological brain tissue in a realistic, draped
surgical environment, the proposed model extends prior VR-based force
metrics and \emph{ex-vivo} simulation approaches into a unified system
that supports objective, biologically grounded assessment of
neurosurgical technical performance.

Protocol 1 showed a strong linear correlation between the sub-skull
sensor and the robotic ground-truth measurements, demonstrating minimal
signal degradation across cranial and soft tissue interfaces. This level
of accuracy is essential to ensure that any detected force reflect
surgical instrument(s) force application rather than system-related
artifacts. This system's ability to maintain a low mean absolute error
under cyclic loading conditions is critical, highlighting its potential
for real-time feedback applications in surgical training.

The sensor's accuracy and minimum detectable force resolution were
confirmed by Protocols 2 and 3, along with establishing a minimum
detectable force threshold of approximately 1 g (9.8 mN). This limit is
within the clinically relevant range for the majority of brain tissue
manipulation although slightly higher than some ultra-sensitive force
sensors used in microneurosurgery \cite{ref3}. This sensor's ability to
detect very small force fluctuations across biologically diverse regions
is critical when utilizing neurosurgical instruments such as the
microscissor and bipolar.

Surgical draping failed to result in any discernible distortion in the
evaluated force signals. This finding is particularly relevant for human
intraoperative translation, as force feedback may otherwise be
attenuated or altered by surgical barriers such as padding and drapes.
Moreover, we demonstrated the stability of force recordings over an
extended timeframe, showing that the system remains reliable even after
prolonged periods of brain tissue manipulation and sustained force
application. To our knowledge, few studies have addressed this issue in
the context of robotic surgical systems. Notably, Hussain et al.
\cite{ref27new} developed a multi-axis capacitive tactile force sensor with
fully decoupled responses to normal and shear forces. The sensor's
repeatability was evaluated through five normal-axis loading cycles with
10-minute intervals, demonstrating good signal stability and a low
repeatability error.

 \clearpage
\begin{figure*}[!t]
    \centering
    \includegraphics[width=0.95\textwidth]{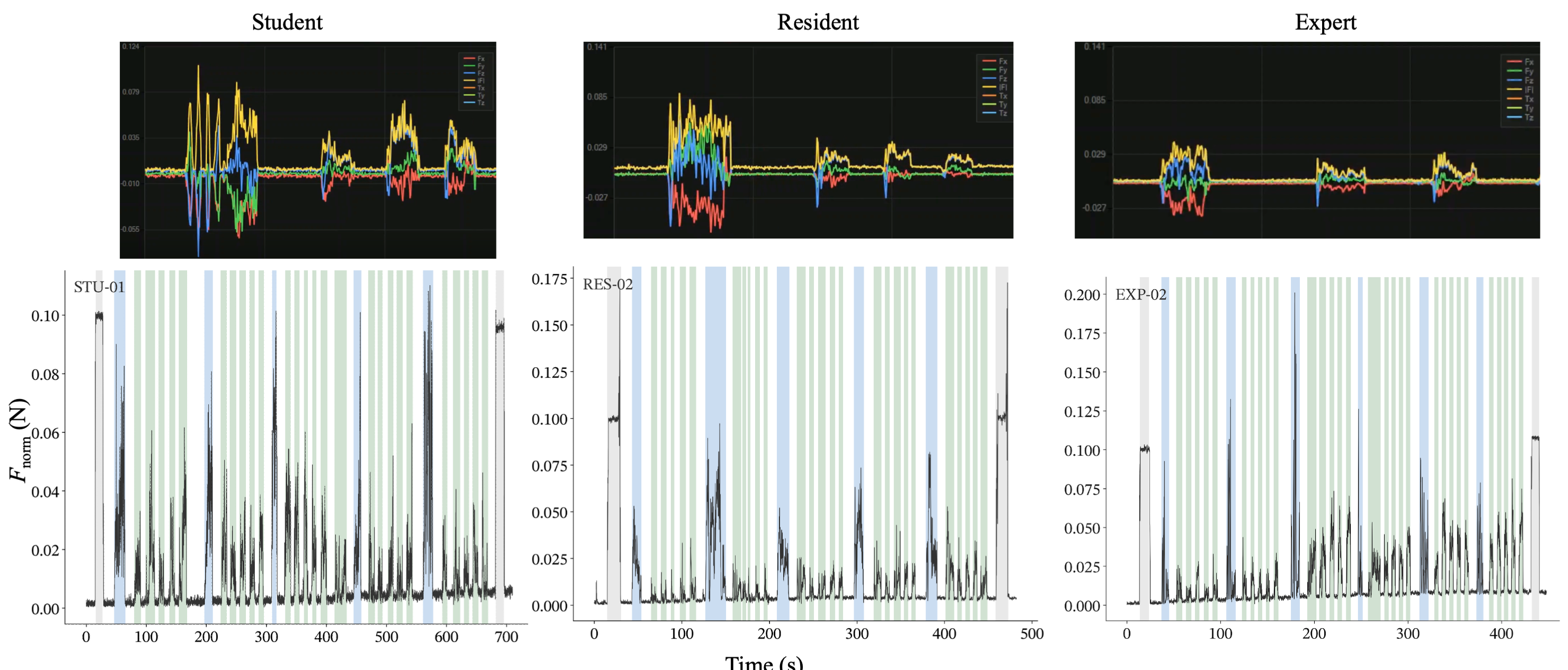}
    \caption{Representative force–time profiles recorded from a 
    student, a resident and an expert neurosurgeon. Top: F norm 
    (yellow) and its individual vector components. Bottom: Time 
    intervals color-coded by instrument activity for metric 
    extraction, where blue indicates microscissor manipulation 
    and green indicates bipolar forceps use.}
    \label{fig:figure7}
\end{figure*}

Errors between protocols were comparable and promising. Protocols 2 and
4 exhibited nearly identical results, with RMSE around 0.0048 N, MRE
around 1.69\%, MaxAE of 0.0095 N, and STD around 0.0029. As expected,
the errors of Protocol 3A were smaller, with an RMSE of 0.0028 N and
MaxAE of 0.0034 N. However, the MRE was slightly higher (1.81\%,
compared to \textasciitilde1.69\%), this may be due to the sensing
element's increased susceptibility to noise at smaller magnitudes of
loading. Nonetheless, these errors are largely well-distributed and
bounded, indicating the repeatability and fidelity of the experimental
setup when compared with reference loading. This is further supported by
the error analysis of all trials, where the linear correlation between
measured and reference values was exceptionally high, at approximately
0.9997, with errors several orders of magnitude smaller than the
reference signals. The linear correlation between the measured signals
and the reference weight application is extremely strong, at around
0.9997, suggesting the overall lack of significant noise/perturbations
in all trials. Furthermore, the mean error across all trials (0.0026 N)
is several orders of magnitude lower than the reference loads applied
(around 100 times larger), further supporting the strong correlation
between measured and reference outcomes. While across all trials, the
errors fluctuate greatly around the mean (STD of 0.0027 N), this
variation is within the same order of magnitude of the mean error,
indicating that the errors are largely bounded.

New advances in surgical force sensing technology are improving feedback
and precision. Fiber-optic and tactile sensors offer small, sensitive
options for real-time force and form detection without sacrificing tool
functionality \cite{ref28new,ref29new}. Although multi-modal sensing haptics-enabled increased accuracy of instrument movements and reduced fatigue, vibration-based feedback aided in controlling applied force and improving training  \cite{ref21new,ref30new}. Intelligent graspers and improved
endoscopic force sensors have also been useful in surgical training
\cite{ref23new,ref24new}. However, there are still problems with workflow
integration, sensor reduction, and real-time data processing. Research
is being done to solve these issues through developments in wireless
transmission, sensor design, and AI integration \cite{ref31new,ref32new,ref33new}.

Our results show the promise of this \emph{ex-vivo} testing environment,
as it can accurately sense the application of reference loads in high
fidelity. Moreover, in applications like neurosurgical training, having
additional telemetry like real-time force application during complicated
procedures can provide novel insights into the way these procedures are
taught. As summarized in Table~\ref{tab:comparison}, prior approaches have achieved either high measurement fidelity through instrument-mounted and robot-integrated sensors, or biological realism through ex vivo simulation, but not both within a single platform. By relocating the sensing element to the tissue side of the interaction, the proposed system measures tool–tissue forces on fresh biological tissue with a resolution of 9.8 mN, without requiring any modification to the surgical instrument. This instrument-agnostic architecture is what allows the same platform to be used with microscissors, bipolar forceps, or any other standard tool, and to be extended to procedures beyond subpial resection. The proposed platform is, to our knowledge, the first to unify biological tissue realism with platform-based, instrument-agnostic sub-skull force sensing, enabling objective measurement of tool-tissue interaction forces during neurosurgical simulation without any modification to the surgical instrument itself.

Calf brain tissue was selected as a neurosurgical surrogate because bovine brains are widely used in simulation studies due to their similar gross neuroanatomy and practical availability. Bovine brain tissue exhibits frequency-dependent viscoelastic behavior, with mean storage and loss moduli of 12.41 kPa and 5.54 kPa, respectively \cite{ref34new}, and shows regional heterogeneity across corpus callosum, corona radiata, cortex, and basal ganglia. While absolute moduli vary widely across species, regions, and loading regimes, these findings are consistent with the known viscoelastic nature of brain tissue.

 Mechanical stability of intact, hydrated bovine brain slices, including both gray and white matter, has been demonstrated for up to five days post mortem under controlled conditions \cite{ref35new}, supporting the reliability of ex vivo testing.

 Species-specific differences nonetheless remain important: the absence of perfusion limits replication of in vivo pressure and turgor conditions, and absolute force magnitudes should not be interpreted as identical to those in living human tissue. These limitations are common to ex vivo surrogate brain models.

 Critically, this platform was not designed to reproduce absolute force magnitudes encountered during live human neurosurgery. Its construct-validity objective is to detect and discriminate relative differences in instrument–tissue interaction patterns across expertise levels, which does not require biomechanical identity between surrogate and human tissue \cite{ref25new}. The relative ranking of force profiles across expert, resident, and student participants remains valid within the biomechanical range of the calf brain model. The platform’s resolution – (detecting loads as low as 1 g (9.8 mN)–falls within the range of forces reported during neurosurgical tissue manipulation.

The platform was designed so that every element except the transducer is user-replaceable: the hemi-skull is regenerated from any CT series through the same open 3D Slicer/Fusion 360 pipeline, allowing craniotomy size, location, and approach to be tailored to a given case or teaching objective; the tissue surrogate can be substituted without altering the acquisition chain; and because the sensing element sits on the tissue side rather than in the instrument, the same setup records tool–tissue force for microscissors, bipolar forceps, aspirators, or any unmodified instrument without recalibration (Figure \ref{fig:figure7}), and accepts alternative ATI Nano17 calibrations for procedures in higher force regimes. This architecture also underlies the system's durability: the transducer is never contacted, handled, or sterilized, so repeated use accumulates only on the low-cost printed skull and mount, and the IP68-rated sensor tolerates irrigation and tissue fluid with overload ratings roughly two orders of magnitude above the $\le 3$ N operating range. Stability under repeated loading is supported directly by the reported protocols, Protocol 5 showed no baseline drift or signal degradation over six continuous loading–unloading cycles (Table VII), Protocols 2 and 3A produced superimposable loading and unloading limbs indicating negligible hysteresis or mechanical settling (Tables III, IV), Protocol 4 confirmed that draping the model for fluid protection costs no measurable accuracy (Table VI), and pooled performance across two specimens, three anatomical locations, and all trials remained consistent (R = 0.9997, RMSE \textless{} 0.005 N, MRE \textless{} 2\%; Fig. 6, Table VIII), with no session-dependent degradation. A formal accelerated-life characterization over extended training use was beyond the scope of this proof-of-concept study and is planned prospectively.

The incorporation of force-sensing devices into surgical training and
assessment is supported by this study. Because the platform is based on
an \emph{ex-vivo} model, it is readily extensible to simulation and
skills evaluation across multiple surgical specialties and procedures.
Such systems may be essential for fostering and benchmarking technical
proficiency in a quantitative and repeatable manner as surgical
education increasingly adopts competency-based, data-driven approaches \cite{ref36new}.

Force sensing has advanced, but it should be improved prior to
implementation into surgical practice. Measurement of accuracy may be
impacted by changes in the characteristics of brain tissue due to aging
or illness \cite{ref37new}. Furthermore, adding force sensors to existing
processes often requires additional equipment or instrument adjustments,
which can be impractical in clinical settings. Widespread adoption is
further limited by insufficient training of surgical teams in the
analysis and interpretation of force data.

The use of two calf brain specimens across the validation protocols reflects a tissue quality-controlled approach. Since the experiments in the study were carried out in 24 hours, the authors believe that very little tissue degradation would occur.  Second, previously experiments carried by Almansouri et al. [\cite{ref9new} in our center have demonstrated both face and content validity.  As this is a proof-of-concept hardware validation study — in which the primary source of variability is instrumentation performance rather than biological heterogeneity — the use of two specimens tested across three independent anatomical locations per brain, yielding between 6 and 18 trials per protocol depending on the experimental design, was considered appropriate to establish sensor accuracy, repeatability, and spatial consistency.

Pilot data from surgeon-performed procedural tasks using microscissors and bipolar forceps across three expertise levels (Figure~\ref{fig:figure7}) provide an early indication of the platform's discriminative potential \cite{ref26new,ref22new}. These experiments are part of a  registered clinical trial (ClinicalTrials.gov, NCT07650253), currently underway, whose primary objective is to determine whether the platform can reliably differentiate force-interaction profiles across distinct neurosurgical instruments, with discrimination of surgical expertise levels as a secondary objective. If supported by the trial's construct validity testing and outcome measures, these metrics could potentially be incorporated into competency-based training curricula to guide deliberate practice. The development of wireless, compact force-sensing systems and their integration with artificial intelligence-based feedback tools represents a promising but as yet unvalidated direction that will require dedicated user studies and clinical trial data before translation into surgical education practice.

Recent advances in artificial intelligence driven surgical education
support the integration of quantitative, real-time feedback systems into
training paradigms. In a randomized clinical trial, Giglio et al.
demonstrated that AI-augmented personalized expert instruction
significantly improved surgical performance, skill transfer, and risk
mitigation compared with either AI or expert feedback alone \cite{ref38new}.
Complementary work has shown that the Virtual Operative Assistant, an
AI-driven system that uses quantitative performance metrics from
surgical simulations to assess skill level and deliver objective,
benchmark-based feedback to trainees, can leverage objective performance
metrics to provide automated, explainable feedback and accurately
differentiate levels of surgical expertise \cite{ref39new}. Similarly,
real-time AI feedback systems have been shown to achieve performance
outcomes comparable to, or exceeding, traditional expert instruction by
delivering continuous, data-driven guidance during simulated procedures
\cite{ref40new}.

However, the implementation of AI-driven curricula is not without
limitations. Prior studies have demonstrated that while AI-enhanced
training can improve safety-related metrics, it may simultaneously
introduce unintended effects on efficiency and movement dynamics,
highlighting the need for careful integration with expert oversight
\cite{ref41new}.

In this context, the integration of real-time force-sensing data into
AI-based tutoring systems represents a natural and necessary evolution.
By combining biomechanical force measurements with intelligent, adaptive
feedback, such systems may enable objective, high-resolution assessment
of surgical performance while preserving the contextual judgment of
expert instructors. This synergy has the potential to advance surgical
education toward fully integrated, data-driven training environments
that optimize both technical precision and learning efficiency. The
development of theses platforms may also result in the development of
intelligent tutoring systems which will enhance training curricula along
with providing surgical educators and learners with transparency and
explainability essential to optimize outcomes.

\section{Conclusion}\label{conclusion}

This work introduces and validates an \emph{ex-vivo} calf-brain surgical
simulation model with a 3D-printed skull and integrated sub-skull force
sensing, capable of capturing applied forces as low as 0.01 N and
providing accurate, reliable measurements of tool-tissue interaction
(including under drapes and during prolonged monitoring) while providing
realistic haptic feedback and an objective framework for assessing
tissue interaction applicable to various surgical procedures.

\section*{Acknowledgements}\label{acknowledgements}

The authors would like to thank Dr. Mojtaba Kheiri of Concordia
University, Montreal, QC, for his support during the experimental phase
of this work, and \'Etienne L\'eger for his assistance in organizing the
laboratory experiments.

\section*{Funding and Disclosures}\label{funding-and-disclosures}

Grant from the Brain Tumour Foundation of Canada, a Medical Education
Research Grant from the Royal College of Physicians and Surgeons of
Canada, the Franco Di Giovanni Foundation, and the Montreal Neurological
Institute and Hospital. Matheus Ballestero was supported by a grant from
FAPESP (S\~ao Paulo Research Foundation, Brazil, No. 2025/10401-0) for a
Research Internship Abroad. Houssem-Eddine Gueziri was supported by
grants from the Canadian Institutes of Health Research (195899). Amir
Hooshiar was supported by Discovery Grant, Natural Science and
Engineering Council of Canada, and CIHR Project Grant.

\end{document}